\documentclass{wscpaperproc}
\usepackage{latexsym}
\usepackage{caption}
\usepackage{subcaption}
\usepackage{graphicx}
\usepackage{mathptmx}
\usepackage[T1]{fontenc}

\usepackage{amsmath}
\usepackage{amsfonts}
\usepackage{amssymb}
\usepackage{amsbsy}
\usepackage{amsthm}
\usepackage{algorithm}
\usepackage{algorithmic}
\usepackage{booktabs}
\usepackage{calc}
\usepackage{xcolor}
\usepackage{graphicx}
\usepackage[mathscr]{euscript}

\usepackage[pdftex,colorlinks=true,urlcolor=blue,citecolor=black,anchorcolor=black,linkcolor=black]{hyperref}

\newtheoremstyle{wsc}
{3pt}
{3pt}
{}
{}
{\bf}
{}
{.5em}
{}

\theoremstyle{wsc}

\begin{document}

\pagestyle{fancyplain}

\thispagestyle{plain}
\firstPageHead{}

\chead{\fancyplain{}{\itshape Inman, Khandait, Pedrielli, and Sankar}}

\rhead{}
\cfoot{}
\renewcommand{\headrulewidth}{0pt} 

\makeatletter
\let\@internalcite\cite
\def\cite{\def\@citeseppen{-1000}%
    \def\@cite##1##2{(##1\if@tempswa , ##2\fi)}%
    \def\citeauthoryear##1##2##3{##1 ##3}\@internalcite}
\def\citeNP{\def\@citeseppen{-1000}%
    \def\@cite##1##2{##1\if@tempswa , ##2\fi}%
    \def\citeauthoryear##1##2##3{##1 ##3}\@internalcite}
\def\citeN{\def\@citeseppen{-1000}%
    \def\@cite##1##2{##1\if@tempswa, ##2)\else{}\fi}%
    \def\citeauthoryear##1##2##3{##1 (##3)}\@citedata}
\def\citeA{\def\@citeseppen{-1000}%
    \def\@cite##1##2{(##1\if@tempswa , ##2\fi)}%
    \def\citeauthoryear##1##2##3{##1}\@internalcite}
\def\citeANP{\def\@citeseppen{-1000}%
    \def\@cite##1##2{##1\if@tempswa , ##2\fi}%
    \def\citeauthoryear##1##2##3{##1}\@internalcite}
\def\shortcite{\def\@citeseppen{-1000}%
    \def\@cite##1##2{(##1\if@tempswa , ##2\fi)}%
    \def\citeauthoryear##1##2##3{##2 ##3}\@internalcite}
\def\shortciteNP{\def\@citeseppen{-1000}%
    \def\@cite##1##2{##1\if@tempswa , ##2\fi}%
    \def\citeauthoryear##1##2##3{##2 ##3}\@internalcite}
\def\shortciteN{\def\@citeseppen{-1000}%
    \def\@cite##1##2{##1\if@tempswa, ##2\else{}\fi}%
    \def\citeauthoryear##1##2##3{##2 (##3)}\@citedata}
\def\shortciteA{\def\@citeseppen{-1000}%
    \def\@cite##1##2{(##1\if@tempswa , ##2\fi)}%
    \def\citeauthoryear##1##2##3{##2}\@internalcite}
\def\shortciteANP{\def\@citeseppen{-1000}%
    \def\@cite##1##2{##1\if@tempswa , ##2\fi}%
    \def\citeauthoryear##1##2##3{##2}\@internalcite}
\def\citeyear{\def\@citeseppen{-1000}%
    \def\@cite##1##2{(##1\if@tempswa , ##2\fi)}%
    \def\citeauthoryear##1##2##3{##3}\@citedata}
\def\citeyearNP{\def\@citeseppen{-1000}%
    \def\@cite##1##2{##1\if@tempswa , ##2\fi}%
    \def\citeauthoryear##1##2##3{##3}\@citedata}
%
%
%
\def\@citedata{%
    \@ifnextchar [{\@tempswatrue\@citedatax}%
                  {\@tempswafalse\@citedatax[]}%
}

\def\@citedatax[#1]#2{%
\if@filesw\immediate\write\@auxout{\string\citation{#2}}\fi%
  \def\@citea{}\@cite{\@for\@citeb:=#2\do%
    {\@citea\def\@citea{, }\@ifundefined
       {b@\@citeb}{{\bf ?}%
       \@warning{Citation `\@citeb' on page \thepage \space undefined}}%
{\csname b@\@citeb\endcsname}}}{#1}}%

%
\def\@citex[#1]#2{%
\if@filesw\immediate\write\@auxout{\string\citation{#2}}\fi%
  \def\@citea{}\@cite{\@for\@citeb:=#2\do%
    {\@citea\def\@citea{; }\@ifundefined
       {b@\@citeb}{{\bf ?}%
       \@warning{Citation `\@citeb' on page \thepage \space undefined}}%
{\csname b@\@citeb\endcsname}}}{#1}}%

%
\def\@biblabel#1{}
\makeatother



\newdimen\bibindent
\bibindent=0.0em
\def\thebibliography#1{\section*{\refname}\list
   {}{\settowidth\labelwidth{[#1]}
   \leftmargin\parindent
   \itemindent -\parindent
   \listparindent \itemindent
   \itemsep 0pt
   \parsep 0pt}
   \def\newblock{}
   \sloppy
   \sfcode`\.=1000\relax}


\setlength{\baselineskip}{12.7pt}

\title{Parameter Optimization with Conscious Allocation (POCA)}

\author{Joshua Inman\\[24pt]
	School of Mathematical Sciences and Statistics\\
	Arizona State University\\
	901 S. Palm Walk\\
	Tempe, AZ 85281, USA\\
\and
Tanmay Khandait\\
Giulia Pedrielli\\[12pt]
School of Computing \& Augmented Intelligence\\
Arizona State University\\
699 S Mill Ave\\
Tempe, AZ 85281, USA\\
\and
Lalitha Sankar\\ [12pt]
School of Computer and Electrical Engineering\\
Arizona State University\\
650 E. Tyler Mall\\
Tempe, AZ 85281, USA\\
}

\maketitle

\section*{ABSTRACT}
The performance of modern machine learning algorithms depends upon the selection of a set of hyperparameters. Common examples of hyperparameters are learning rate and the number of layers in a dense neural network. Auto-ML is a branch of optimization that has produced important contributions in this area. Within Auto-ML, hyperband-based approaches, which eliminate poorly-performing configurations after evaluating them at low budgets, are among the most effective. However, the performance of these algorithms strongly depends on how effectively they allocate the computational budget to various hyperparameter configurations.  
We present the new Parameter Optimization with Conscious Allocation (POCA), a hyperband-based algorithm that adaptively allocates the inputted budget to the hyperparameter configurations it generates following a Bayesian sampling scheme. 
We compare POCA to its nearest competitor at optimizing the hyperparameters of an artificial toy function and a deep neural network and find that POCA 
finds strong configurations faster in both settings.

\section{Introduction}\label{sec:intro}
Hyperparamaters play a crucial role in the performance of machine learning models. However, properly tuning such hyperparameters is often a costly procedure, particularly in settings with a large number of hyperparameters. As a result, there has been increasing interest in developing efficient hyperparameter tuning algorithms, and several algorithmic approaches have been explored. One family of approaches, as proposed in~\shortcite{JMLR:v13:bergstra12a} and others, is to use a Bayesian model to sequentially select hyperparameter configurations (henceforth referred to as simply ``configurations''). Another family of approaches, including \textit{Successive Halving}~\shortcite{pmlr-v51-jamieson16} and \textit{Hyperband}~\shortcite{JMLR:v18:16-558}, samples a set of configurations but repeatedly discards the worst performing configurations, running the remaining configurations to a larger budget. Significantly improved performance is found by algorithms, such as Bayesian Optimization and Hyperband (BOHB)~\shortcite{pmlr-v80-falkner18a}, that combine these approaches by selecting configurations via a Bayesian model and then using Hyperband to decide how long to test each configuration. However, this more nuanced approach still struggles in the face of a large hyperparameter configuration space, which is difficult to efficiently search through with a limited budget. An approach that more wisely allocates the overall budget of the hyperparameter optimization (HPO) process is a promising area for potential improvement. In this paper, we present for the first time the algorithm Parameter Optimization with Conscious Allocation (POCA), an algorithm for hyperparameter optimization, which seeks to improve on existing methods by allocating less of the overall budget to configurations found early in the experiment (which are either selected uniformly at random or by a Bayesian model with little data), allowing more of the overall budget to be used on more promising configurations, while also being more explorative of the hyperparameter configuration space at the start of the HPO process.

In the remainder of the paper, Section~\ref{sec:sota} introduces the relevant literature on existing hyperparameter tuning algorithms. Section \ref{sec:algoOver} introduces POCA. Finally, in Section \ref{Sec: Experiments}, we show POCA's performance on an artificial toy function and on a machine learning dataset against its nearest competitor BOHB~\shortcite{pmlr-v80-falkner18a}.

\section{Related Work}\label{sec:sota}
A variety of approaches for hyperparameter optimization have been proposed. In general, algorithms fall into at least one of three categories: model-free, Bayesian Optimization, and multi-fidelity approaches.

\subsection{Model-free Algorithms}\label{sec:mfapproach}
Many of the earliest approaches were model-free approaches. Grid-search~\shortcite{montgomery2017design}, for example, creates a set of configurations by taking the cross product of a discretization of each hyperparameter domain and then runs each configuration to a fixed budget. This approach becomes infeasible when there are a large number of hyperparameters. An alternate approach is random search~\shortcite{JMLR:v13:bergstra12a}, which randomly selects configurations and runs them to a fixed budget. It is more feasible than grid search and has become a natural baseline for HPO algorithms.

Other model-free approaches include population-based methods, which maintain a population of configurations and then construct a new generation of improved configurations by mutating and combining configurations. In particular, Differential Evolution~\shortcite{10.5555/2765832} has been used as a component of the multi-fidelity algorithm Differential Evolution and Hyperband (DEHB), which we discuss in Section~\ref{sec:combo_approaches}.

\subsection{Bayesian Optimization}\label{sec:BOapproach}
A second approach to HPO is Bayesian optimization. Bayesian optimization involves two parts: (i) a probabilistic surrogate model, which is fitted to all the existing datapoints, as and when new data becomes available, and (ii) an acquisition function, which is used to select new configurations. Many Bayesian optimization approaches use the expected improvement as the acquisition function; this allows selecting the configuration with the highest expectation of being an improvement over the previous best-found configuration. However, performance depends strongly on the choice of surrogate. 
 A traditional surrogate are Gaussian processes (GP)~\shortcite{10.7551/mitpress/3206.001.0001}, which have the advantage of being fully specified by their mean and covariance functions. However, GPs scale cubically in the number of data points being sampled and also do not scale well to high-dimensions, posing challenges for HPO problems. This has led other surrogates to be explored. One such surrogate is Random Forests~\shortcite{10.1007/978-3-642-25566-3_40}, which benefit from being able to handle large and complex hyperparameter configuration spaces naturally and without the scaling problems of GPs. Another approach is using Tree Parzen Estimators (TPE)~\shortcite{10.5555/2986459.2986743}. Given a $\gamma$-percentile loss $y^*$, the TPE models the densities $l(\bold{\lambda}) = p(\bold{\lambda} | y < y^*)$ and $g(\bold{\lambda})=p(\bold{\lambda} | y \ge y^*)$ instead of $p(y | \bold{\lambda})$, where $\bold{\lambda}$ is a configuration and $y$ is a possible loss value. 
\shortciteN{pmlr-v162-song22b} showed that maximizing the probability of improvement is equivalent to maximizing $\frac{l(\bold{\lambda})}{g(\bold{\lambda})}$.

\subsection{Multi-fidelity Algorithms}\label{sec:multifapproach}
A third approach to HPO is multi-fidelity algorithms, which acquire low budget (or fidelity) approximations of the performance of a configuration and then decide whether to evaluate the configuration further. Learning-based curve prediction algorithms, such as those discussed in~\shortcite{KOHAVI1995304} and~\shortcite{10.1145/312129.312188}, model the improvement of a configuration as it is given more resources and terminates it if it is predicted that it will not perform better than the best-found configuration. Bandit-based multi-fidelity algorithms have met particular success. In this class, Successive Halving (SH)~\shortcite{pmlr-v51-jamieson16} selects a set of configurations, runs them to a certain fixed budget, and then terminates the worst-performing proportion $\eta$, running the remaining configuration to $\frac{1}{\eta}$-times the budget until only one configuration remains. While Successive Halving shows promise, it creates a difficult tradeoff between the initial fixed budget and the number of configurations to test. Hyperband~\shortcite{JMLR:v18:16-558} runs a collection of Successive Halving brackets, each with different settings for the initial fixed budget. Because of this, Hyperband can hedge its bets against the tradeoff that Successive Halving faces. 

\subsection{Hybrid Approaches}
\label{sec:combo_approaches}
Hyperband is still limited by having no mechanism to use existing data to improve its performance. In an attempt to fix this challenge, it has been combined with several Bayesian optimization or model-free approaches to fix this flaw. Some examples include BOHB~\shortcite{pmlr-v80-falkner18a}, which uses a TPE to select new configurations for new hyperbands; Sequential Model-Based Algorithm Configuration (SMAC) \shortcite{JMLR:v23:21-0888}, which uses a Random Forest in a similar manner; and DEHB~\shortcite{ijcai2021p296}, which mixes Hyperband and Differential Evolution. Other multi-fidelity approaches select the budgets to which to train configurations actively, unlike the previous approaches which construct a pre-defined schedule. Multi-task Bayesian optimization~\shortcite{10.5555/2999792.2999836} models multiple related tasks simultaneously. It begins by searching the configuration space of the cheaper task before switching to the more expensive task later in the HPO process. Given the interest in HPO in recent years, the full extent of the literature is vast and for reasons of space, cannot be thoroughly covered. For a more detailed view of the landscape, we refer to~\shortcite{Feurer2019}.

\subsection{Problem Statement}\label{sec::probstate}
Following the notation in~\shortcite{Feurer2019}, we denote a machine learning algorithm as $\mathscr{A}$. Each $\mathscr{A}$ has $N$ hyperparameters, each with a domain $\Delta_i$. Each $\Delta_i$ may be real-valued, integer-valued, binary, or categorical and may include conditionality, that is, a hyperparameter may only be used by $\mathscr{A}$ if a different hyperparameter is assigned a certain value. The hyperparameter configuration space is denoted $\Delta = \Delta_1 \times \Delta_2 \times \ldots \times \Delta_N$. A hyperparameter configuration $\bold{\lambda} \in \Delta$ is a vector of hyperparameters, and $\mathscr{A}$ running with a given hyperparameter configuration $\bold{\lambda}$ is written as $\mathscr{A}_{\bold{\lambda}}$.

Then, given a dataset $\mathscr{D}$ with a corresponding training dataset $D_{train}$ and validation dataset $D_{valid}$, the goal of hyperparameter optimization is find the best possible hyperparameter configuration, as given by the following minimization problem:
\begin{equation}
    \bold{\lambda}^* = \text{argmin}_{\bold{\lambda} \in \Lambda} \mathbb{E}_{\mathscr{D} | (D_{train}, D_{valid})} f(A_\bold{\lambda}, D_{train}, D_{valid}), \label{eq:lambda-opt}
\end{equation}

where $f$, the validation protocol, is a function that measures the loss of a model generated by $A_{\bold{\lambda}}$ trained on $D_{train}$ and evaluated on $D_{valid}$. Misclassification rate is a common validation protocol; we denote this quantity by $\bold{\lambda}^*$, defined in \eqref{eq:lambda-opt}, and note that it almost always must be estimated. We write $\hat{\bold{\lambda}}^*$ to denote an estimate of $\bold{\lambda}^*$.

\vspace{3pt}
\framebox{\parbox{0.9\linewidth}{\textbf{Example}. Consider the HPO problem of tuning the following hyperparameters of a neural network: batch size, learning rate, max dropout, momentum, and weight decay. Each hyperparameter would have an associated domain, e.g. learning rate may be restricted to [0.1, 0.0001], and the Cartesian Product of these ranges forms the hyperparameter configuration space. It the goal of the HPO algorithm to find a configuration $\bold{\lambda}^*$ such that if a neural network was trained for every configuration in the configuration space, the neural network configured with $\bold{\lambda}^*$ would have the maximum test accuracy.}}
\vspace{3pt}

\subsection{Hyperband}
A hyperband \shortcite{JMLR:v18:16-558} helps provide a solution to the HPO problem by running a sequence of successive halving brackets (described in Section~\ref{sec:multifapproach}). Each successive halving bracket runs a collection of configurations up to a certain budget, eliminates the worst-performing configurations based upon the performance obtained with the allocated budget, and then evaluates the remaining configurations further to a larger budget. Successive halving itself contains hyperparameters, such as the initial number of configurations to evaluate and the initial budget to evaluate those configurations prior to the first halving point. Hyperband runs a sequence of successive halving brackets, each having different settings for the aforementioned hyperparameters. In particular, more \textit{explorative} settings will correspond to hyperbands with a large number of initial configurations while more \textit{exploitative} settings will be elicited by hyperbands with large initial budgets. 

Once all the successive halving brackets have been executed, the configuration that performs the best is selected as $\hat{\bold{\lambda}}^*$. Algorithms with hyperband backbones have emerged as the dominant approach for solving the HPO problem~\shortcite{JMLR:v23:21-0888,ijcai2021p296}.

\subsection{Contributions}
The contributions of POCA beyond existing Hyperband-based Bayesian HPO algorithms are as follows:
\begin{enumerate}
    \item POCA exploits the improvement of the Bayesian model throughout the training process by selecting configurations from the Bayesian model with a increasing probability through the training process.
    \item POCA further exploits the Bayesian model's improvement by allocating more of the total budget on hyperbands that occur later in the experiment. POCA accomplishes this by shortening the hyperbands that occur earlier in the experiment, which feature configurations that were primarily selected uniformly at random or by a Bayesian model with limited data available. In addition, this permits POCA to be more explorative at the start of the HPO process.
    \item In contrast to BOHB, POCA allows the surrogate model to use all existing datapoints rather than only allowing data at the highest budget. This gives the Bayesian optimizer more information to model the surrogate of the loss across the hyperparameter configuration space, and it helps prevent the selection of a previously-evaluated configuration which was known to be poor.
\end{enumerate} 
These contributions can be extremely relevant for complex HPO problems where the vast majority of the hyperparameter configuration space will not lead to good performance. In such instances, the configurations that are selected uniformly at random or by a Bayesian model trained with limited data at the start of an HPO process are likely to be poor choices. POCA will be beneficial in such cases as it allows for exploration of a wider proportion of the configuration space early in the HPO process. POCA allocates a reduced budget to the poor configurations that are likely at the outset, thereby saving resources to explore the more promising configurations found by the Bayesian model when more data becomes available.

\section{PARAMETER OPTIMIZATION with CONSCIOUS ALLOCATION}\label{sec:algoOver}
\begin{figure}[H]
\centering
\includegraphics[width=\textwidth]{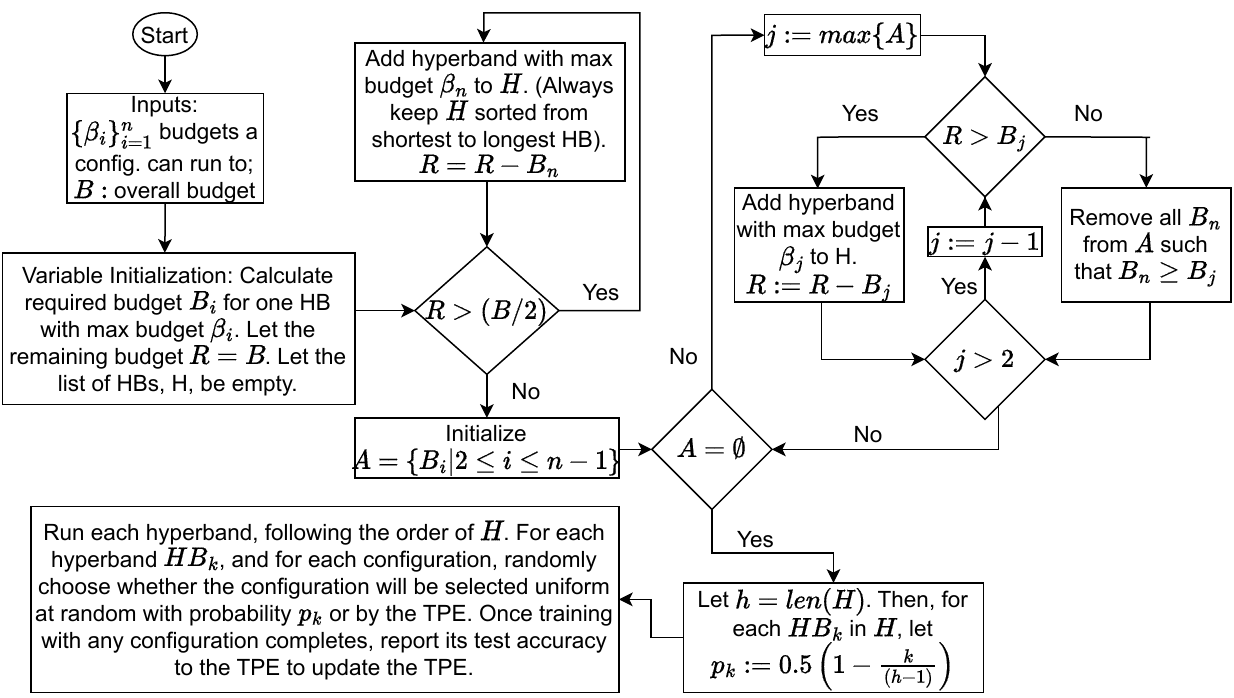}
\caption{POCA flowchart.} 
\label{Image: Flowchart}
\end{figure}

In this section, we provide a detailed description of the POCA algorithm, and discuss the intuition behind the algorithm. POCA executes a sequence of hyperbands, selecting a proportion of configurations using a Tree Parzen Estimator (TPE). It consists of three primary components: the budget allocation, the configuration selection, and the surrogate update. In brief, the budget allocation procedure determines how to allocate the total budget for the experiment between hyperbands of different lengths, where the length of a hyperband is defined as the largest budget to which a configuration may run. The configuration selection determines whether a newly-selected configuration should be selected uniformly at random or by the TPE. Lastly, the surrogate update provides new data to the TPE whenever a configuration reaches a halving point within a Successive Halving iteration. Figure \ref{Image: Flowchart} shows a flowchart of the algorithm.

The budget allocation procedure, described in detail in Section~\ref{Budget Allocation}, takes the total budget for the entire HPO process and a list of budgets to which a single hyperparameter configuration may be tested, and constructs the sequence of hyperbands that will run. The sequence of hyperbands will have the following intrinsic properties:
\begin{enumerate}
\item The sequence of hyperbands will be of non-decreasing length.
\item At least half of the total budget will be used on hyperbands of maximum length. This ensures that the optimization procedure has sufficient resources to evaluate configurations at the budget with the high accuracy estimate of their performance (e.g., validation loss).
\item The difference between the total budget allowed and the total budget effectively used by the hyperband schedule is less than the budget required by a hyperband with the smallest possible length.
\end{enumerate}
 The intuition behind POCA, illustrated in Figure \ref{fig: Budget Allocation}, relies on the assumption that the surrogate improves as it receives more data. Under this assumption, the configurations generated by the TPE later in the experiment are generally of higher quality than the configurations generated earlier in the experiment. The budget allocation procedure exploits this fact by limiting the length of the hyperbands run at the start of the HPO process, which in turn limits the budget assigned to the generally poor configurations generated at the start of the process. Instead, this budget is used to explore more configurations overall, which allows more data to be delivered to the TPE. Under the assumption, this will then improve the quality of the configurations generated by the TPE later in the process. Further, by sampling more configurations early in the process, POCA is able to find better configurations early in the experiment simply by sampling more configurations. The configuration selection procedure is likewise designed to trust the TPE more as it receives more data. Hence, it selects a low proportion of configurations from the TPE at the start of the HPO process but gradually increases that proportion as the HPO process proceeds.

The hyperbands are then run in ascending order by length, and the surrogate, the TPE, is updated whenever a configuration has been trained to a halving point in its Successive Halving bracket. The TPE, described in detail by Section~\ref{Surrogate Update}, consists of only one pair of kernel densities, in contrast to~\shortcite{pmlr-v80-falkner18a}, which used one pair of kernel densities for every configuration budget. This ensures that each collected data point is considered by the TPE whenever it selects a new configuration.

\subsection{Hyperband}\label{sec:hyperB} 

Figure~\ref{fig: POCA Budget} shows an example of a schedule for hyperbands generated by POCA. We give as input the total POCA budget $T$ ($T=600$ in Figure~\ref{fig: POCA Budget}), a minimum configuration budget $b_{\mbox{\tiny{min}}}$ ($b_{\mbox{\tiny{min}}}=1$ in Figure~\ref{fig: POCA Budget}), a maximum configuration budget $b_{\mbox{\tiny{max}}}$ ($b_{\mbox{\tiny{max}}}=8$ in Figure~\ref{fig: POCA Budget}), and a promotion rate $\eta$ ($\eta=0.5$ in Figure~\ref{fig: POCA Budget}) defined as the proportion of configurations that will be promoted to a larger budget at each halving point within a SH bracket.

Iteratively, hyperband $HB_k$ will schedule $NSH_k$ successive halving brackets, each with index $s_{j,k}, j=1,\ldots,NSH_k;k=1,\ldots,h$. Each successive halving bracket will be assigned a number of configurations $c_{j,k}, j=1,\ldots,NSH_k;k=1,\ldots,h$. Of these, a proportion $p_k$ is sampled uniformly at random, while $1-p_k$ is selected according to the TPE surrogate and the associated acquisition function. Each configuration is evaluated with minimum budget of $b_{\mbox{\tiny{min}},j,k}$ and at each successive halving iterate $i=1,\ldots,\lfloor \log_{1/\eta}b_k \rfloor + 1$, it is increased by an amount $b_{min}/\eta^i, i=1,\ldots,\lfloor\log_{1/\eta}b_k \rfloor + 1$, where $b_k$ is the maximum budget for the $k$-th hyperband. In the following subsections, we explain the hyperband schedule in terms of how to determine the number of hyperbands to run and the maximum budget to allocate to each hyperband.

\subsection{Budget Allocation}\label{Budget Allocation}
Given the total budget $T$ (in terms of number of epochs, training time, etc.), the promotion rate $\eta \in (0,1)$, and the maximum budget $b_{\mbox{\tiny{max}}}$,  we determine the number of hyperbands and the maximum budget allowed to train configurations for the $k$-th hyperband, i.e., $b_k$. 

As a first step, we define a sequence of maximum budgets $\left\lbrace \beta_i \right\rbrace, i=2,\ldots,n$, ($\beta_1=b_{\mbox{\tiny{min}}}$ and it is an input to POCA). Note that there is generally a lower number of maximum budgets than the possible number of hyperbands. Equivalently, multiple hyperbands can use the same maximum budget, i.e., $b_k = \beta_i$ for multiple values of $k$. Define a candidate hyperband $P_i$ characterized by $i-1$ successive halving brackets (where $i-1$ is a candidate value for $NSH$ associated to the candidate hyperband) with a total budget $B_i = (\lfloor \log_{1/\eta}(b_i) \rfloor + 1) b_i$. Given $NSH,B_i, \beta_i$ for each SH, with $j=1,\ldots,NSH$, and each hyperband $HB_k$ scheduled from the $i$-th candidate, the number of configurations $c_{jk}$ is derived using the approach in~\shortcite{JMLR:v18:16-558}. We will refer to the set of candidate hyperbands as $P=\left\lbrace P_i\right\rbrace^{n}_{i=2}$, and the set of total budgets associated to the candidate hyperbands as $A=\left\lbrace B_i\right\rbrace^{n}_{i=2}$. 

We are left with defining the hyperband scheduling. This activity determines how many hyperbands POCA will execute for each candidate before generating a new one, and so on until the budget is exhausted. The scheduling starts with available budget $R$ equal to the total budget $T$, with an empty list $H$ of scheduled hyperbands. We will refer to $h$ as the size of the list $H$ and initialize it to $0$, and we will use $k$ to iterate the elements of the list $H$, i.e., $k = 1, \ldots, h, h>0$.  We start scheduling hyperbands with the largest possible budget. In particular, we keep inserting hyperbands with maximum budget $b_{\mbox{\tiny{max}}}$ and a total associated budget $B_n$ until half of the total budget is exhausted, i.e., $R < \frac{T}{2}$. At this point, we proceed by adding to the list hyperbands from the set of candidates $P_i,i=n-1,\ldots,2$ (from the \textit{largest} to the \textit{smallest}) and updating $R$ by subtracting the associated $B_i$. If at any step a candidate becomes infeasible ($R<B_{i}$), the candidate with the closest feasible budget is added instead. We keep following this sequence until no candidate hyperband can be run with the remaining budget, i.e., $R > B_2$.

\begin{figure}[h]
\centering
\begin{subfigure}[t]{.35\textwidth}
    \centering
    \raisebox{2.5cm}{\includegraphics[width=\textwidth]{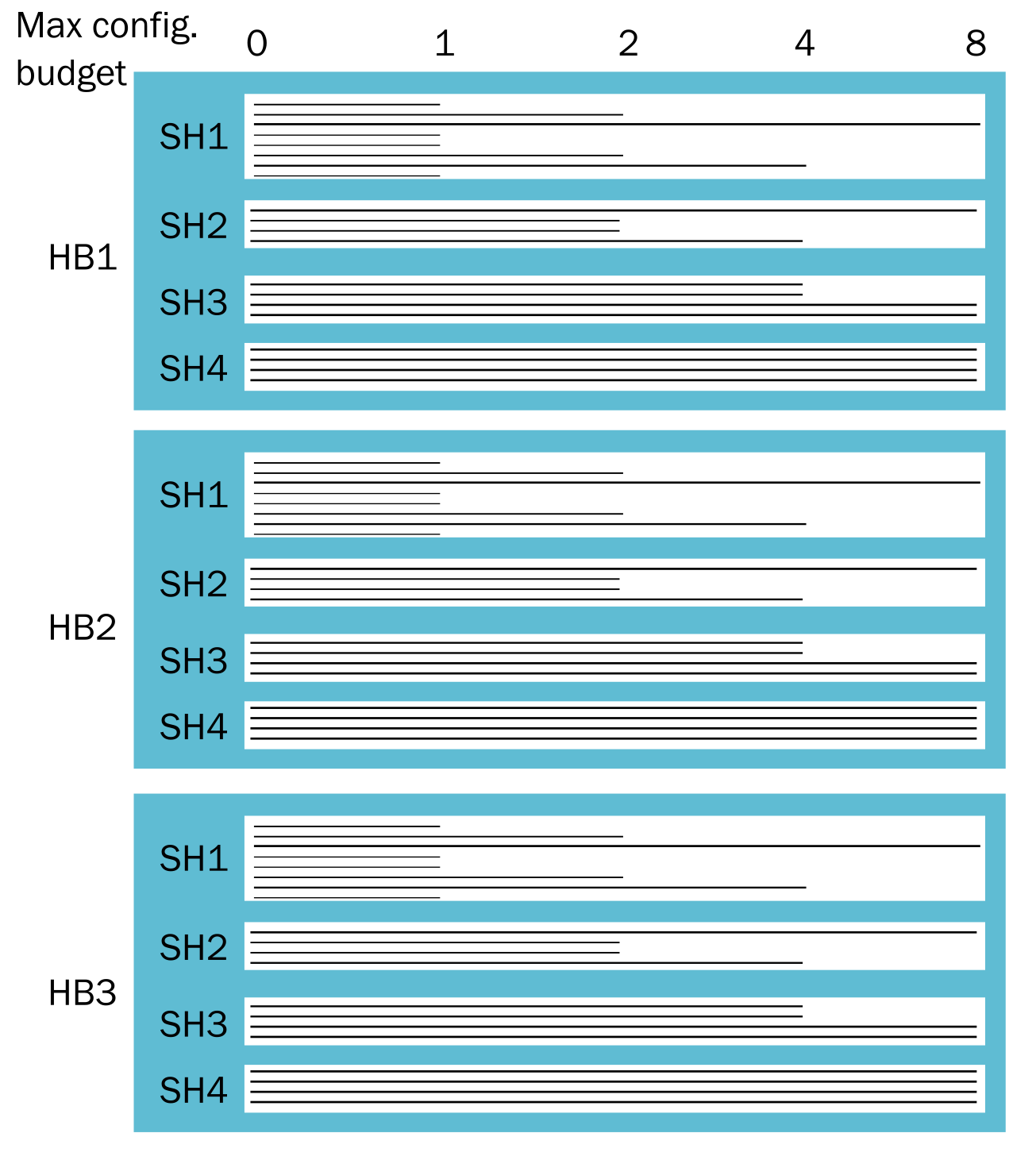}}
    \caption{BOHB}
\end{subfigure}
\begin{subfigure}[t]{.35\textwidth}
   \centering
   \includegraphics[width=\textwidth]{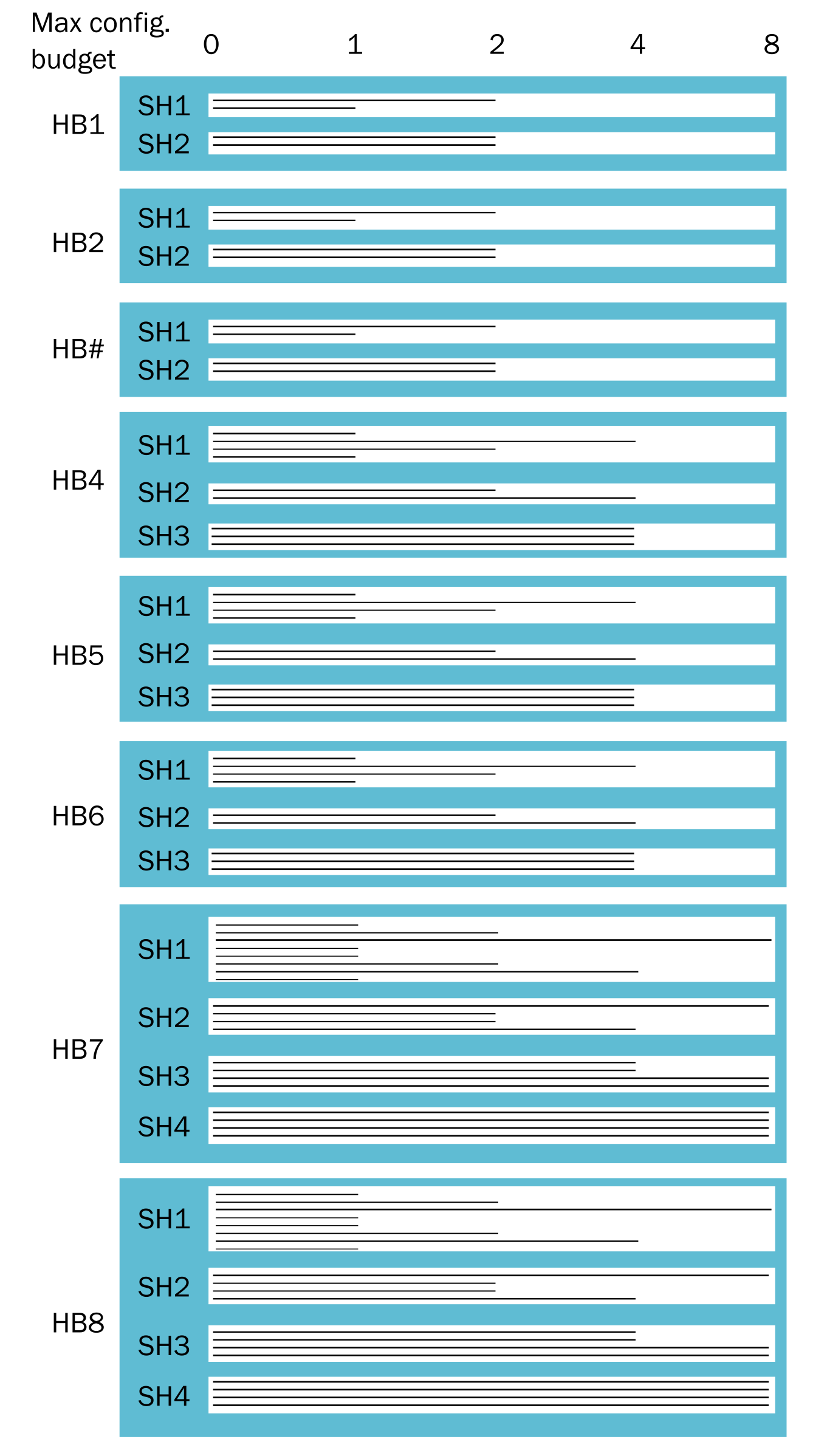}
   \caption{POCA}
   \label{fig: POCA Budget}
\end{subfigure}
\caption{Hyperbands scheduled by BOHB and POCA for the example in Section \ref{Budget Allocation}. Each horizontal bar represents a single configuration.} \label{fig: Budget Allocation}
\end{figure}

In the following, we show an example (represented in Figure~\ref{fig: Budget Allocation}), while the pseudocode for the generation of the candidate hyperbands and the schedule is provided in Algorithm~\ref{Scheduling_Algorithm}. 

\vspace{3pt}
\fbox{\parbox{0.9\linewidth}{\textbf{Example. }Consider a case where the total budget available is $T=600$, with the budget sequence $\left\lbrace \beta_i\right\rbrace$ set to $\left\lbrace 1, 2, 4, 8\right\rbrace$. Hyperband or BOHB would run $5$ hyperbands, each with a maximum per-configuration budget of $8$. This would lead to a total of $100$ configurations being sampled. On the other hand, POCA would run $6$ hyperbands with maximum per-configuration budget of $2$, another $6$ with maximum per-configuration budget of $4$, and then $3$ configurations with a maximum per-configuration budget of $8$. This allows POCA to test $138$ configurations total, by training less configurations to the maximum number of epochs.}}
\vspace{3pt}

\begin{algorithm}[H]
    \caption{POCA: Budget Allocation}
    \label{Scheduling_Algorithm}

    \begin{algorithmic}
    \STATE {\bfseries Input: } Total budget $T$, list of candidate hyperbands $\{P_i\}_{i=2}^n$ with corresponding hyperband budgets $\{B_i\}_{i=2}^n$.

    \STATE {\bfseries Output: } $H$, a list of scheduled hyperbands of length $h$;

    \vspace{2pt}
    \hrule 
    \vspace{2pt}

    \STATE \textbf{Initialization: } Set the remaining budget $R = T$. Set the number of scheduled hyperbands $h=0$, and initialize the list of scheduled hyperbands $H$ to be empty. Initalize $A = \{B_i | 2 \le i \le n-1\}$;
    \WHILE{$R < \frac{T}{2}$}
        \STATE Append hyperband $P_n$ to $H$. Update $h \leftarrow h + 1$, and $R \leftarrow R-B_n$;
    \ENDWHILE
    \WHILE {$A \neq \emptyset$}
        \FOR{$j \in [\max\{A\}, 2]$}
            \IF {$B_j < R$}
                \STATE Append hyperband $P_j$ to $H$. Update $h \leftarrow h + 1$, and $R \leftarrow R-B_j$;
            \ELSE
                \STATE Remove $B_j$ and all $B_k > B_j$ from $A$;
            \ENDIF
        \ENDFOR
    \ENDWHILE
    \STATE return $H$
    \end{algorithmic}
\end{algorithm}

\subsection{Adaptive Configuration Selection}
\label{Adaptive Configuration Selection}
This sequence of hyperbands generated by the Budget Allocation procedure is then an input in the configuration selection procedure. All the configurations are initially selected uniformly at random until there is sufficient data to build the first TPE surrogate. As soon as the surrogate becomes available, each configuration is randomly generated by the TPE-associated acquisition function with probability $1-p_k$, or by the uniform random selection procedure with probability $p_k$. The parameter $p_k$ is specific to each hyperband $HB_k$. For the first hyperband, it is set to $0.5$, and for the final hyperband, it is set to $0.0$. For the hyperbands in between, $p_k$ declines linearly between these two constants (so $p_k = 0.5(1-\frac{k}{h-1})$ for all $k$). This is a preliminary schedule that we propose in this paper, but more complex schedules, including cases where the $p$ remains away from $0$, are part of current and future work (see Section~\ref{sec:concl}).

\vspace{3pt}
\framebox{\parbox{0.9\linewidth}{\textbf{Example}. For an experiment with only three hyperbands, BOHB would assign $p=0.33$ for each configuration of each hyperband. However, POCA would assign $p=0.5$ for configurations of the first hyperband, $p=0.25$ for those of the second hyperband, and $p=0.0$ for configurations in the final hyperband.}}
\vspace{3pt}

\subsection{Surrogate Update}\label{Surrogate Update}
POCA uses a Tree Parzen Estimator (TPE) as its surrogate model.
The construction of the TPE~\shortcite{10.5555/2986459.2986743} requires a percentile $\gamma$ and previously evaluated configurations $\bold{\lambda}_1, \ldots, \bold{\lambda}_n$ with corresponding validation losses $y_1, \dots, y_n$ to be given. Then, the $\gamma$-percentile loss $y^*$ is determined. Given a loss function $f$, it then creates two multi-dimensional kernel density estimators (KDEs), $l(\bold{\lambda})$ and $g(\bold{\lambda})$, where $l(\bold{\lambda})$ is formed by the observations $\{\bold{\lambda}^{(i)}\}$ where $f(\bold{\lambda}^{(i)}) < y^*$, and $g(\bold{\lambda})$ is formed using the other observations. In both $l(\bold{\lambda})$ and $g(\bold{\lambda})$, continuous hyperparameters have a Gaussian kernel, and categorical hyperparameters have an Aitchison-Aitken kernel. Following~\shortcite{JMLR:v13:bergstra12a}, we used the KDE implementation from Statsmodels~\shortcite{Seabold2010StatsmodelsEA}, and we adopted Scott's rule for efficient bandwidth estimation. Unlike~\shortcite{JMLR:v13:bergstra12a}, which created different densities for each budget level (and then only uses the densities generated on the largest budgets), POCA creates only one pair of densities, using each available datapoint to construct it. This allows all previously discovered information to be used in generating new configurations. 

\vspace{3pt}
\framebox{\parbox{0.9\linewidth}{\textbf{Example}. Suppose enough configurations have been tested to a budget of $b$ and $3b$ to form kernel densities at both budgets. Then, three configurations were run to budget $b$, all of which had similar performance. However, only one of these configurations is \textit{promoted} to the budget of $3b$. In the TPE used in BOHB, only the configuration that was promoted will inform the new configurations selection. On the contrary, in POCA, all three configurations would be considered.}}
\vspace{3pt}

\noindent Finally, to generate a new configuration, the TPE draws a large number of configurations and returns the configuration $\sigma$ that maximizes $\frac{g(\bold{\lambda})}{l(\bold{\lambda})}$, which \shortcite{pmlr-v162-song22b} showed was equivalent to maximizing the probability of improvement.

\section{Numerical Analysis}
\label{Sec: Experiments}
In this section, we provide experimental results of the performance of POCA in a variety of settings, compared to BOHB~\shortcite{pmlr-v80-falkner18a}. The code for POCA has been available at this address: www.github.com/SankarLab/Parameter-Optimization-with-Conscious-Allocation-POCA-.

\subsection{Artificial Toy Function: Counting Ones}
In this experiment, we investigated the performance of POCA on an artificial toy function that features a high-dimensional hyperparameter space and a mix of categorical and continuous hyperparameters. Define a sequence of $N_{cat}$ categorical variables $x_i \in \{0,1\}$ from $i = 1,\ldots,N_{cat}$ and a sequence of $N_{cont}$ continuous variables $y_j \in [0,1]$ with $j=1, \ldots, N_{cont}$. Further, define a sequence of Bernoulli Random Variables $L_j$ with parameter $y_j$ and $j=1,\ldots,N_{cont}$ and a vector containing an instance of all of these variables as $\bold{x}=[x_1, \ldots,x_{N_{cat}},y_1, \ldots, y_{N_{cont}}]$. Then, we define a minimization problem known as the Counting Ones problem as:

    $$f(\bold{x}) = \sum_{i=1}^{N_{cont}} x_i + \sum_{j=1}^{N_{cat}} \mathbb{E}[L_j].$$

In this scenario, the budget is defined as the number of samples drawn to estimate the expectation for each $L_j$. The simple problem provides insight in how HPO algorithms perform in high-dimension spaces with a mix of continuous and categorical variables.

\begin{figure}
    \centering
    \includegraphics[width = 0.45\textwidth]{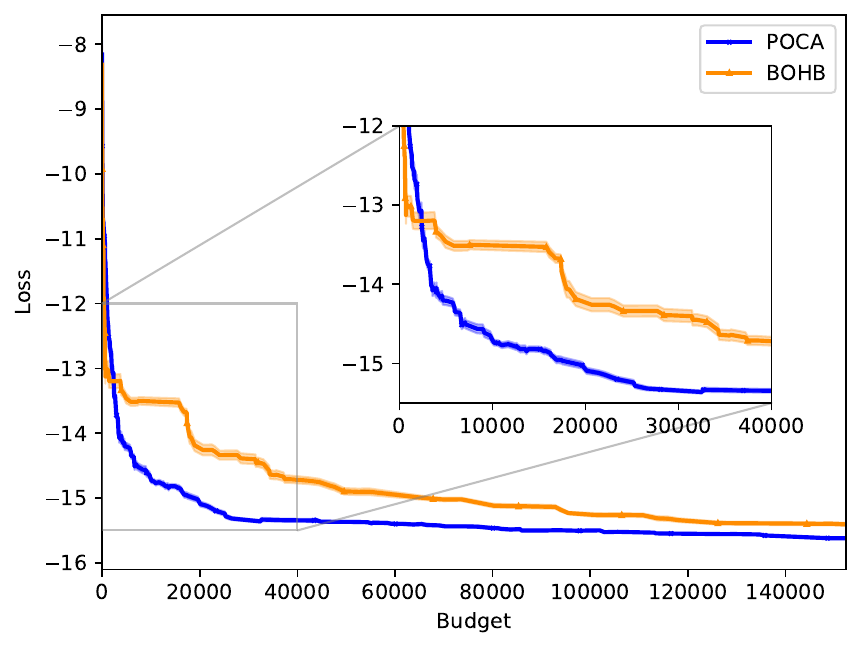}
    \caption{Average loss of the best configuration found by POCA and BOHB for the Counting Ones experiment with 95\% confidence intervals.}
    \label{fig:exp_countingones}
\end{figure}

This experiment was run with $b_{\mbox{\tiny{min}}} = 9$, $b_{\mbox{\tiny{max}}}=720$, $T=153,100$ , and $\eta = \frac{1}{3}$. Thus, the possible configuration budgets for both BOHB and POCA was as follows: $\beta_1=9, \beta_2=27, \beta_3 = 81, \beta_4 = 243,\text{ and, }\beta_5 = 729$. We provide the results in Table~\ref{tab:cones_exp} and Figure~\ref{fig:exp_countingones} over $100$ replications.

The results show that after the full budget, POCA  on average finds a configuration with true loss of -15.753, and BOHB finds an average true loss of -15.428. The difference is statistically significant with 95\% confidence. Further, Figure \ref{fig:exp_countingones} shows that POCA finds strong configurations at much lower overall budget than BOHB does, and POCA's best-found configuration after $40,000$ samples is comparable to BOHB's best-found configuration at the full budget. Numerical results are shown in Table \ref{tab:cones_exp}.

\begin{table}[htbp]
\centering
\caption{Result for Counting Ones Experiment. Here we report the mean loss for BOHB and POCA, along with the standard error, and the lower and upper confidence bound at $95\%$ confidence.}
\label{tab:cones_exp}
\begin{tabular}{@{}lllll@{}}
\toprule
Experiment & Mean Loss & Std. Error & LCB & UCB \\ \midrule
BOHB & -15.428 & 0.036 & -15.499 & -15.357 \\
POCA & -15.753 & 0.013 & -15.779 & -15.727 \\ \bottomrule
\end{tabular}%
\end{table}

\subsection{MNIST}
In this experiment, POCA is tested on a Convolutional Neural Network model on a subset of the MNIST dataset consisting of $8{,}192$ points in the training set, $1{,}024$ points in validation set, and $10{,}000$ points in the testing set. For the model, we optimize nine hyperparameters related to the architecture of the neural network. This experiment was run with $b_{\mbox{\tiny{min}}} = 9$, $b_{\mbox{\tiny{max}}}=720$, $T=30{,}650$, and $\eta = \frac{1}{3}$. Thus, the possible configuration budgets for both POCA and BOHB was as follows: $\beta_1=9, \beta_2=27, \beta_3 = 81, \beta_4 = 243,\text{ and, }\beta_5 = 729$. 

For this experiment, there were nine hyperparameters to tune. The following five hyperparameters had integer domains: number of convolutional (conv.) layers ([1,3]), number of filters in the first conv. layer ([4, 64]), the number of filters in the second conv. layer ([4, 64]), the number of filters in the third conv. layer ([4, 64]), and the number of hidden units in the fully connected layer ([8, 256]). The following three hyperparameters had domains that were intervals of the real numbers: the learning rate ([1e-6, 1e-2]), the SGD momentum ([0, 0.99]), and the dropout rate ([0, 0.9]). The final hyperparameter was the optimizer of the model weights, which was a categorical choice between the Adam or the SGD optimizer. The SGD momentum hyperparameter was conditional on the optimizer being SGD, and the hyperparameters for the number of filters in the second and third conv. layer was conditional on that layer existing.

\begin{figure}[htbp]
     \centering
     \begin{subfigure}[b]{0.45\textwidth}
         \centering
         \includegraphics[width=\textwidth]{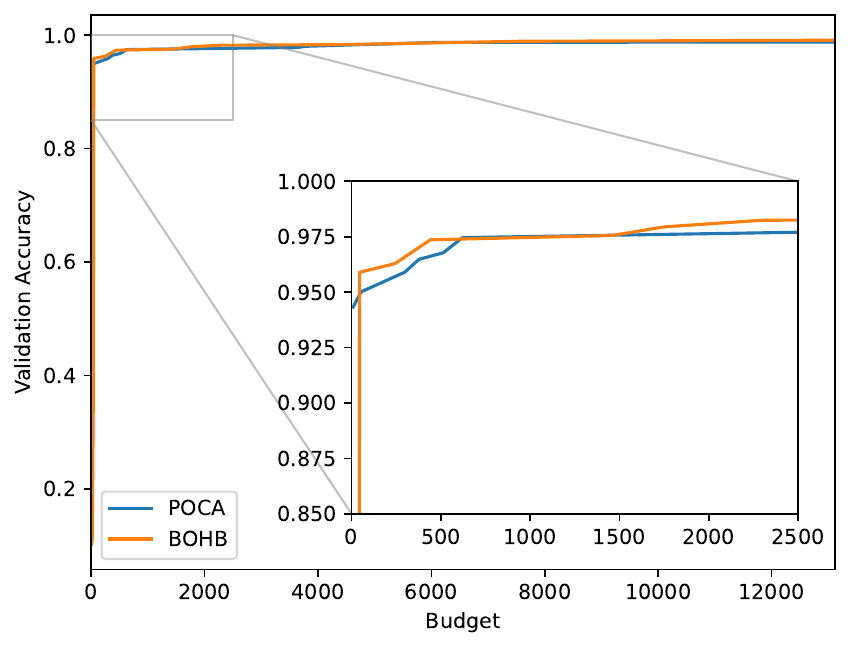}
         \caption{Validation Accuracy}
         \label{fig:mnist_exp1_vali}
     \end{subfigure}
     \hfill
     \begin{subfigure}[b]{0.45\textwidth}
         \centering
         \includegraphics[width=\textwidth]{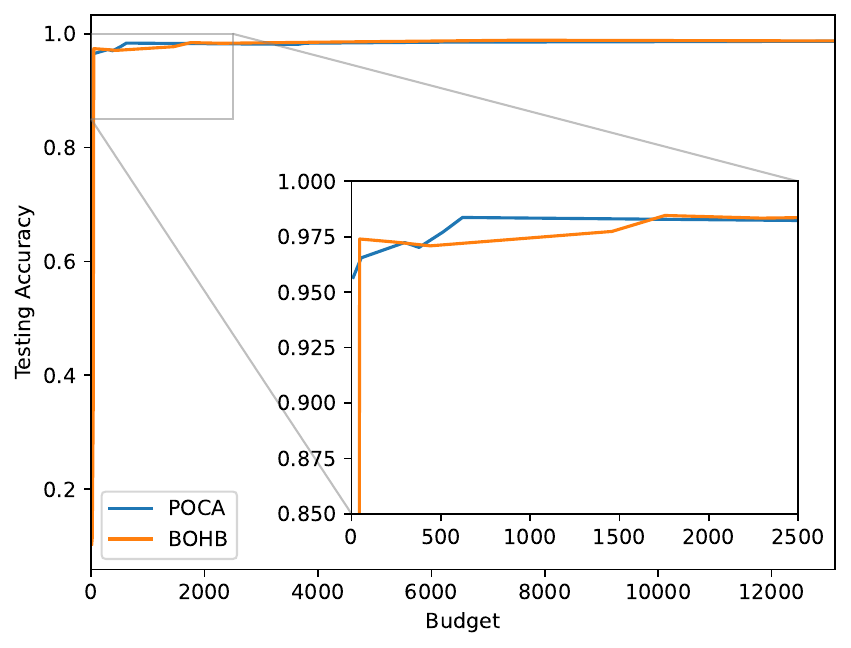}
         \caption{Test Accuracy}
         \label{fig:mnist_exp1_test}
     \end{subfigure}
        \caption{Accuracy plots for MNIST.}
        \label{fig:mnist_exp1}
\end{figure}

The results of the experiment are shown in Figure \ref{fig:mnist_exp1}. At each epoch, the graphs show the best-found accuracy by any previously tested configuration, regardless of what budget was used to test that configuration. 
For this experiment, POCA and BOHB perform similarly by the end of the HPO process, but POCA finds strong configurations about 1,000 epochs prior to BOHB.

\section{Conclusions}\label{sec:concl}
In this paper, we presented a parameter optimization algorithm POCA, which proposes a new way to adaptively allocate an increasing budget to hyperbands that are sequentially scheduled. This allows the algorithm to be more explorative of a wide-range of configurations at the start of the HPO process and gives the Bayesian model more information to choose new configurations later in the experiment. On the artificial toy function Counting Ones, POCA found strong configurations quicker and found a better final configuration than BOHB. On an MNIST experiment, POCA found adequate configurations earlier in the HPO process than BOHB and found similarly-effective final configurations. 

Concerning future developments in POCA, the difference between the total budget allowed and the total budget effectively used by the Hyperband schedule is less than the budget required by a Hyperband with the smallest possible length. In this sense, according to the current implementation we may not exhaust the budget entirely. Current research is exploring alternative formulations for the full utilization of the budget. Another aspect relates the choice of the sequence of sampling configurations either uniformly or from the TPE. Further work will relate the sequence to the performance of the hyperbands and possibly consider more than two schemes. In addition, the current TPE utilized by POCA does not account for the noise caused by evaluating configurations to different budgets. Further research will explore Bayesian optimizers that are capable of adequetely modeling this noise, and expand on the range of the experiments.

\section{Acknowledgements}
This work is supported in part by NSF grants \#2046588, \#2134256, \#1815361, \#2031799, \#2205080, and \#1901243 and by DARPA ARCOS program under contract \#FA8750-20-C-0507, and Lockheed Martin funded contract \#FA8750-22-9-0001.

\footnotesize

\bibliographystyle{wsc}

\bibliography{POCA}

\section*{AUTHOR BIOGRAPHIES}

\noindent {\bf JOSHUA INMAN} is an undergraduate student pursuing concurrent Honors degrees in Mathematics (Statistics) and Computer Science at Arizona State University. His research interests include the theoretical foundations of machine learning including enhanced algorithms for hyperparameter optimization. His email is jainman2@asu.edu.\\

\noindent {\bf TANMAY KHANDAIT} is a Ph.D candidate and graduate research assistant in School of Computing and Augmented Intelligence at ASU.
He graduated with Masters in Computer Science from Arizona State University. 
His research interests include machine learning, verification of cyber-physical systems, computer vision, and 
bilevel optimization. His email is tkhandai@asu.edu.\\

\noindent {\bf GIULIA PEDRIELLI} received her Master and Ph.D in Mechanical Engineering from Politecnico di Milano, Italy, in 2009 and 2013, respectively.
She is currently an Associate Professor for the School of Computing and Augmented Intelligence (SCAI) at Arizona State University. Her research activity is in the field of stochastic simulation and optimization with focus on Bayesian Optimization.
She serves as Associated Editor for the Journal of Simulation and the Journal of Flexible Service and Manufacturing. Her email is gpedriel@asu.edu.\\

\noindent {\bf LALITHA SANKAR} received her Ph.D. in Electrical Engineering from Rutgers University.
She is currently a Professor at Arizona State University. Her research interests lie at the intersection of information theory and machine learning and its applications to complex networks including the electric power grid. She serves as an Associate Editor for the IEEE Transactions on Information Forensics and Security, as well as the IEEE Information Theory Society BITS magazine. Her email is lsankar@asu.edu.

\end{document}